# Learning representations in Bayesian Confidence Propagation neural networks

Naresh Balaji Ravichandran
Computational Brain Science Lab
KTH Royal Institute of Technology
Stockholm, Sweden
nbrav@kth.se

Anders Lansner
Computational Brain Science Lab
Stockholm University and KTH Royal
Institute of Technology
Stockholm, Sweden
ala@kth.se

Pawel Herman
Computational Brain Science Lab
KTH Royal Institute of Technology
Stockholm, Sweden
paherman@kth.se

*Abstract*—Unsupervised learning of hierarchical representations has been one of the most vibrant research directions in deep learning during recent years. In this work we study biologically inspired unsupervised strategies in neural networks based on local Hebbian learning. We propose new mechanisms to extend the Bayesian Confidence Propagating Neural Network (BCPNN) architecture, and demonstrate their capability for unsupervised learning of salient hidden representations when tested on the MNIST dataset.

*Keywords—neural networks, bio-inspired, brain-like, unsupervised learning, structural plasticity.*

## I. Introduction

Artificial neural networks (ANN) have made remarkable progress in supervised pattern recognition in recent years. ANNs achieve this mainly under the umbrella of deep learning by discovering hierarchies of abstract features in the data using multiple layers of distributed representations. At this stage, it is valuable to study how they compare with the biological neural networks, and explore new opportunities at this intersection.

We see at least three fundamental differences between current deep learning approaches and the brain:

Firstly, most deep learning methods rely extensively on labelled samples for learning the entire hierarchy of representations, although biological systems mostly learn in an unsupervised fashion. Recent work in deep learning research has increasingly paid attention to developing unsupervised learning methods [1, 2, 3], and the work we present here will also be in this direction.

Secondly, deep learning methods predominantly make use of error back-propagation (backprop) for learning the weights in the network. Although extremely efficient, backprop has several issues that make it an unlikely candidate model for synaptic plasticity in the brain. The most apparent issue is that the synaptic connection strength between two biological neurons is expected to comply with Hebb's postulate, i.e. to depend only on the available local information provided by the activities of the pre- and postsynaptic neurons. This is violated in backprop, since synaptic weight updates need gradient signals to be communicated from distant output layers. We refer to other work for a detailed review and possible biologically plausible alternatives to backprop [4].

Thirdly, an important difference between current deep ANNs and the brain concerns the abundance of recurrent connections in the latter. A typical cortical area receives on the order of 10% of synapses from lower order structures, e.g. thalamus, and the rest from other cortical neurons [5]. In contrast, deep learning networks rely predominantly on feed-forward connectivity. The surplus 90% connections are likely involved in associative memory, constraint-satisfaction, top-down modulation and selective attention [5]. However, we will not consider those important aspects of cortical computation in this work.

This motivates exploring alternative more biologically plausible learning strategies that enable unsupervised learning of representations using local Hebbian rules. The approach we follow here involves framing the update and learning steps of the neural network as probabilistic computations. Probabilistic approaches are widely used in both deep learning models [3] and computational models of brain function [6]. One disadvantage of probabilistic models is that the known methods do not scale well in practice. Also, inference and learning with distributed representations is often intractable and we invariably resort to approximation [3, 7].

In this work, we further adopt a modular network architecture used for more biologically detailed cortical memory models [7, 8] and earlier abstract work [9, 10]. The networks are modularized in terms of hypercolumns comprising a number of functional minicolumns interacting in a soft-winner-take-all manner. The abstract view of a hypercolumn is that it represents some attribute, e.g. edge orientation, in a discrete coded manner. A minicolumn unit represents a local subnetwork of around a hundred recurrently connected neurons with similar receptive field properties. Such an architecture was initially generalized from primary visual cortex, but today has more support also from later experimental work and has featured in spiking computational models of cortex [11, 12, 13].

## II. Related work

A popular variety of unsupervised learning approach is to train a hidden layer to reproduce the input data, for example,

autoencoders and Restricted Boltzmann Machines (RBM). The autoencoder and RBM networks trained with just one hidden layer are relevant here since the learning of weights of the connections from the input to hidden layers rely on local gradients, and the representations can be stacked on top of each other to learn hierarchical features. However, stacked autoencoders and stacked RBMs are only used as pre-training procedures on which end-to-end supervised fine-tuning (using backprop) is performed [2]. Other unsupervised methods like variational autoencoders and generative adversarial networks are very promising, but they generally depend on training the network with backprop.

Recent work by Krotov and Hopfield [14] addresses this specific problem by learning hidden representations solely using an unsupervised method. Their network trains the input to hidden feed-forward connections along with additional (non-plastic) recurrent inhibitory connections that provides competition within the hidden layer. For evaluating the representation, the weights are frozen, and another layer connecting the labels is trained using a linear classifier. Our approach shares some common features with that of Krotov and Hopfield [14], e.g. learning hidden representations by unsupervised methods, and evaluating the representations by another supervised classifier. However, this work differs by following a probabilistic approach extending the BCPNN architecture (explained in the next section).

All the related models we have discussed so far employ either recurrent connectivity within the hidden layer, or hidden-to-input feed-back connections, or both. In this work, we only use feed-forward connections, along with an implicit competition via a local softmax operation.

It is also observed that, for unsupervised learning, having sparse connectivity in the feed-forward connections performs better than full connectivity [15]. The unsupervised learning methods we have discussed so far, however, employ full connectivity [14, 15]. In addition to the unsupervised methods, networks employing supervised learning like convolutional neural networks (CNNs) force a fixed spatial filter to obtain this sparse connectivity. Here we take an alternate approach where, along with learning the weights of the feed-forward connections, which is regarded as biological synaptic plasticity, we also simultaneously learn the sparse connectivity between the input and hidden layer, in analogy with the structural plasticity in the brain [16].

### III. BAYESIAN CONFIDENCE PROPAGATION NEURAL NETWORK

We describe the network architecture and update rules for the Bayesian Confidence Propagation Neural Network (BCPNN). The simplest BCPNN architecture for classification contains two layers, one for data and the other for labels.

A layer consists of a set of hypercolumns (HC), each of which represents a discrete random variable $X_i$ (upper case). Each HC, in turn, is composed of a set of minicolumns (MC) representing a particular instance of the random variable $x_i$ (lower case). The probability of the variable $X_i$ is then a multinomial distribution, defined as $p(X_i = x_i)$, such that $\sum_{x_i} p(X_i = x_i) = 1$. In the neural network, the activity of the MC is interpreted as $p(X_i = x_i)$, and the sum of activities of all the MCs inside a HC sums to one.

Since the network is a probabilistic graphical model, we can compute the posterior of a target HC in the label layer conditioned on all the source HCs in the input layer. We will use $x$'s and $y$'s for referring the HCs in the input and output layer respectively. Computing the exact posterior $p(Y_j|X_{1:N})$ over the target HC is intractable, since it scales exponentially with the number of units. The assumptions $p(X_1,..,X_N|Y_j) = \prod_{i=1}^{N} p(X_i|Y_j)$ and $p(X_1,..,X_N) = \prod_{i=1}^{N} p(X_i)$ allows us to write the posterior as:

$$p(Y_j|X_1,..,X_N) = p(Y_j) \frac{p(X_1,..,X_N|Y_j)}{p(X_1,..,X_N)}$$

$$= p(Y_j) \prod_{i=1}^{N} \frac{p(X_i|Y_j)}{p(X_i)}$$

$$= p(Y_j) \prod_{i=1}^{N} \frac{p(X_i,Y_j)}{p(X_i)p(Y_j)}$$

When the network is driven by input data $\{X_1,..,X_N\} = \{x_1^D,..,x_N^D\}$, we can write the posterior probabilities of a target MC in terms of the source MCs as:

$$p(y_j|x_1^D,..,x_N^D) = p(y_j) \prod_{i=1}^{N} \frac{p(x_i^D,y_j)}{p(x_i^D)p(y_j)}$$

$$= p(y_j) \prod_{i=1}^{N} \prod_{x_i} \left( \frac{p(x_i,y_j)}{p(x_i)p(y_j)} \right)^{I(x_i = x_i^D)}$$

where $I(\cdot)$ is the indicator function that equals 1 if its argument is true, and zero otherwise. We have written the posterior of the target MC as a function of all the source MCs (all $x_i$'s). The log posterior can be written as:

$$log\ p(y_j|x_1^D,..,x_N^D) = log\ p(y_j) + \sum_{i=1}^{N} \sum_{x_i} I(x_i = x_i^D)\ log\ \frac{p(x_i,y_j)}{p(x_i)p(y_j)}$$

Since the posterior is linear in the indicator function of data sample, $I(x_i = x_i^D)$ can be approximated by its expected value, that is, $p(x_i^D)$. Except for $p(x_i^D)$, all the terms in the posterior are functions of the marginals $p(x_i)$, $p(x_j)$, and $p(x_i, x_j)$. We define the terms bias $\beta(y_j) = log\ p(y_j)$ and weight $w(x_i, y_j) = log\ \frac{p(x_i,y_j)}{p(x_i)p(y_j)}$ in analogy with artificial neural networks.

The inference step to calculate the posterior probabilities of the target MCs conditioned on the input sample is given by the activity update equations:

$$h(y_j) = \beta(y_j) + \sum_{i=1}^{N} \sum_{x_i} p(x_i^D)\ w(x_i, y_j)$$

$$\pi(y_j) = \frac{exp(\gamma h(y_j))}{\sum_k exp(\gamma h(y_k))} \quad (1)$$

where $h(y_j)$ is the total input received by each target MC from which the activity $\pi(y_j)$ is recovered by softmax normalization (with gain $\gamma$) within the HC.

The learning step involves incrementally updating all the marginals as input samples are presented. The marginals, bias and weight parameters are updated as follows:

$$\begin{aligned}
\tau_p \frac{dp(x_i)}{dt} &= k_p \left(\pi(x_i) - p(x_i)\right), \\
\tau_p \frac{dp(x_i, y_j)}{dt} &= k_p \left(\pi(x_i)\pi(y_j) - p(x_i, y_j)\right), \\
\tau_p \frac{dp(y_j)}{dt} &= k_p \left(\pi(y_j) - p(y_j)\right), \\
\beta(y_j) &= k_\beta \log p(y_j), \\
w(x_i, y_j) &= k_w \log \frac{p(x_i, y_j)}{p(x_i)p(y_j)}.
\end{aligned} \quad (2)$$

The terms $k_p$, $k_\beta$, $k_w$, and $\tau_p$ are the plasticity gain, bias gain, weight gain, and learning time constant, respectively. Equations 1 and 2 define the complete set of update and learning equations of the BCPNN architecture.

The scope of the work is limited to this abstract model of BCPNN where MCs are the fundamental computational unit. The use of this network architecture and learning rule to model short- and long-term memory with palimpsest properties in biological memory context can be found in our previous work [6, 17].

## IV. Unsupervised representation learning

The network for unsupervised learning is similar to the two-layer network, except we now have more than one HC, each of which can contain arbitrary number of MCs (see Figure 2). On top of this network, we introduce additional mechanisms that enable learning representations.

### A. Bias regulation

The BCPNN update rule implements Bayesian inference if the parameters are learnt with the source and target layer probabilities available as observations. When the target layer is hidden, we are learning the representations, and we cannot expect the update rule to follow Bayesian inference. In fact, we can see that performing learning and inference simultaneously is counter-productive in this case.

Consider a hidden representation with a noisy initialization that assigns some MCs with slightly higher marginal probability $p(z_j)$ than others. Learning will then amplify this difference, and find parameters that will associate more input samples with the MCs with high $p(z_j)$, causing the marginals to increase further. One way to circumvent this effect is to "push" MCs that have low $p(z_j)$ to be more active in the future, a kind of activity homeostasis in biological terms

We use a bias regulation mechanism, where the bias gain $k_\beta$ for each MC, which equals 1 if we are performing just Bayesian inference, is made a function of $p(z_j)$. One motivation for choosing the bias gain is that, we want to influence the marginal $p(z_j)$ alone, leaving out the weight parameters that is responsible for learning the input to hidden mapping. The value of $p(z_j)$ is compared with respect to the maximum entropy probability, $p_{MaxEnt} = 1/N_{MC}$, where $N_{MC}$ is the number of MCs in a HC. Notice that the maximum entropy distribution is the ideal representation without the input layer, since all the MCs have equal marginal probability, and hence $p_{MaxEnt}$ acts as the reference for bias regulation. The update equation of $k_\beta$ is:

$$\tau_k \frac{dk_\beta}{dt} = 1 + (k_\beta - 1)\frac{(p_{MaxEnt}/4)^2}{(p(z_j) - p_{MaxEnt}/4)^2}$$

where $\tau_k$ is the time constant. The mechanism maintains the value of gain $k_\beta$ at around one when $p(z_j) \gg p_{MaxEnt}$, and drops sharply to negative values when $p(z_j)$ is below $p_{MaxEnt}$ (see Figure 1). The rate of this drop is controlled using the parameter $k_{half}$, defined as the value of gain $k_\beta = k_{half}$ at $p(z_j) = 1/2\, p_{MaxEnt}$.

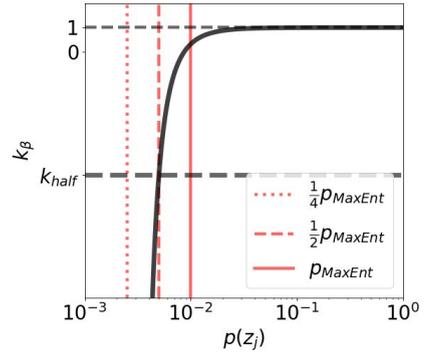

Figure 1: Bias regulation mechanism. For generating the figure, $k_{half} = -5$ and $p_{MaxEnt} = 0.01$ was used.

### B. Structural plasticity

Structural plasticity aims to find receptive fields for the hidden MCs from the input layer. We define a boolean variable $M_{ij}$ denoting the connection from the $i$ th input HC to the $j$ th hidden HC as either active $M_{ij} = 1$ or silent $M_{ij} = 0$. Each $M_{ij}$ is initialized randomly with probability $p_M$. Once initialized, the total number of active incoming connections to each hidden HC is fixed whereas the outgoing connections from a source HC can be changed. The mutual information (MI) between the $i$ th HC and $j$ th HC is calculated from the BCPNN weights: $I_{ij} = \sum_{x_i, z_j} p(x_i, z_j) w(x_i, z_j)$. The input HCs then normalizes the MI by the total number of active outgoing connections:

$$\widehat{I}_{ij} = I_{ij} / (1 + \sum_k M_{ik}).$$

Since the total number of active incoming connections is fixed, each hidden HC greedily maximizes the $\widehat{I}_{ij}$ it receives by removing the active connection with the lowest $\widehat{I}_{ij}$ (set $M_{ij}$ from 1 to 0) and add an inactive connection with the highest $\widehat{I}_{ij}$ (set $M_{ij}$ from 0 to 1). We call this operation a *flip*, and use a parameter $N_{flips}$ to set the number of flips made per training iteration.

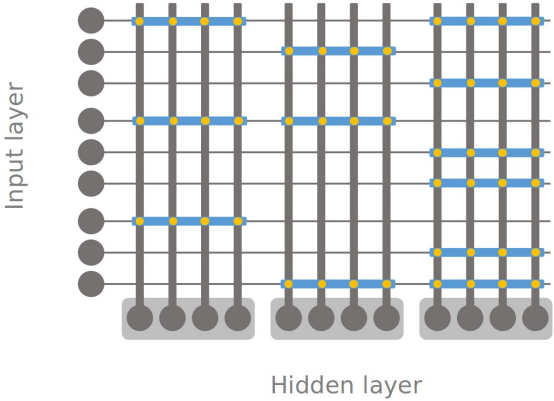

Figure 2: The schematic of the network used for unsupervised learning. In this network, the input layer contains nine binary HCs (grey circles on the left), and the hidden layer contains three HCs (grey boxes), each of which contains four MCs (grey circles inside the boxes). The existence of a connection between an input HC and hidden HC is shown as a blue strip, i.e., $M_{ij} = 1$. The input-hidden weights are shown as yellow dots and are present only when a connection already exists.

## V. CLASSIFICATION

After learning the input-hidden connection, we freeze the weights, biases, and receptive fields of this connection, and treat the hidden layer representations as input to train a BCPNN classifier with an output (label) layer. We add another BCPNN projection from hidden to output layer with a negative gain $k_w = -1$ (in contrast to the existing projection with $k_w = 1$). This is analogous to a network architecture used to model reinforcement learning in the basal ganglia [18]. We call the projections *Go* ($k_w = 1$) and *No-Go* ($k_w = -1$), as they are intended to increase the probability of correct labels and reduce the probability of wrong labels, respectively. The classification layer training procedure is as follows: we first drive the pre-trained network from input samples and check the predicted label. If the classification is wrong, we either train the Go projection (by setting the output activations to the true label), or train the No-Go projection (by setting the output activations to the predicted label), or both. We run this procedure for $N_{sup}$ epochs.

## VI. RESULTS

We evaluate the model using the MNIST hand-written digits database [19]. The MNIST dataset contains 60000 training and 10000 test 28x28 images. The grey-scale intensities were normalized to the range [0,1] and interpreted as probabilities. For each of the following subsections, we used 50000 random training samples for training, report on the other 10000 in the validation set, and at the end of this section, report the test accuracy of 10000 samples for the best set of model parameters. The network had 784 input HCs, the hidden layer had 30 HCs and 100 MCs per HC, and the output layer had one HC with 10 MCs corresponding to digit labels. The time constants $\tau_k$ and $\tau_p$ were scaled by the training time, and we varied these scaling factors $\tau_k^o$ and $\tau_p^o$ in the experiments. The parameters used in the simulation are listed in Table 1. All the results presented here are the mean and standard deviation of the mean squared validation error over 10 random runs of the network, unless stated otherwise.

TABLE I. MODEL PARAMETERS

| Symbol | Value | Description |
|---|---|---|
| $\Delta t$ | 0.01 | Time-step |
| $\gamma$ | 1 | Softmax gain |
| $k_{half}$ | -100 | Bias gain when marginal is $1/2\, p_{MaxEnt}$ |
| $\tau_p^o$ | 0.5 | Multiplier for learning time-constant |
| $\tau_k^o$ | 0.1 | Multiplier for bias gain time-constant |
| $p_M$ | 0.1 | Probability of connections from input to hidden layer |
| $N_{train}, N_{val}$ | 50k, 10k | Number of training and validation samples |
| $N_{usup}$ | 5 | Number of epochs of training for unsupervised learning |
| $N_{sup}$ | 25 | Number of epochs for training the BCPNN classifier |

### A. Bias regulation

We evaluate the bias regulation mechanism by measuring the accuracy while varying the relevant parameters: $k_{half}$ is changed from $-10$ to $-100$ in steps of $-10$, and $\tau_k^o$ from $10^{-2}$ to $10^1$ in exponential steps of $10$. Results are shown in Figure 3a. The accuracy improves consistently as $k_{half}$ is lowered and converges at around $k_{half} < -50$ to $96.7$. This suggests that our bias regulation mechanism effectively improves the representations.

To quantitatively assess the effect of $k_{half}$ on the marginals $p(z_j)$, we compute the marginal entropy of each the HC, $H(p(Z_j)) = \sum_{z_j} p(z_j)\, log\, p(z_j)$, and plot the histogram of this entropy over the 30 HCs (Figure 3c). Note that for the marginals, higher entropy is preferred since it indicates all the MCs in a HC are utilized evenly, and $p_{MaxEnt}$ was our target while designing the bias regulation mechanism. Even though the number of samples used in plotting this histogram is low ($N_{HC}$), it clearly shows that lowering the $k_{half}$ increases the entropy of all the HCs.

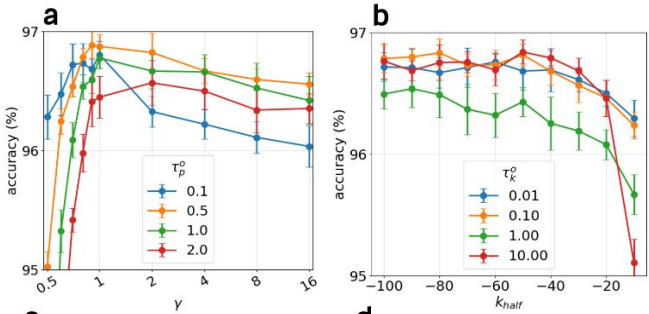

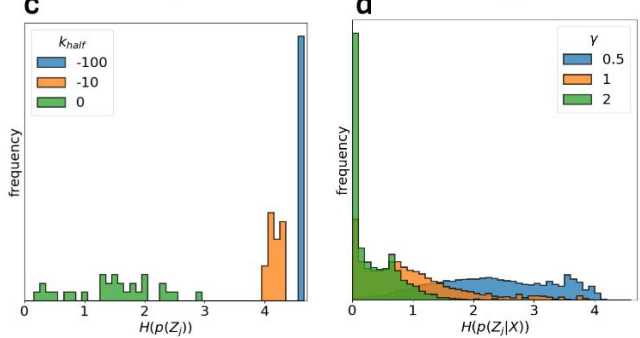

Figure 3: Accuracy results as a function of $k_{half}$ for different values of time constant of bias gain $\tau_k^o$ (3a), accuracy results as a function of softmax gain $\gamma$ for different values of learning time constant (3b). Histogram of marginal entropy $p(Z_j)$ of hidden layer HCs for different $k_{half}$ (3c), and histogram of conditional entropy $p(Z_j|X)$ for different values of softmax gain $\gamma$ (3d).

The marginals give the overall utilization of the MCs over the training set, and we have evaluated it by measuring the entropy of this marginal distribution with respect to the parameter that regulates the bias ($k_{half}$). However, the marginals by themselves cannot give the complete picture of the representations since they do not take into account how well the representations differentiate between samples. To see this, we can imagine a worse-case scenario where all the MCs, for all the input samples, have a posterior of $p_{MaxEnt}$. This will result in high marginal entropy, but is certainly not desirable.

We measured the entropy of the posterior distribution of the MCs conditioned on each input sample, that is, $H(p(Z_j|X))$. Contrary to the entropy of the marginals, we expect this entropy to be as low as possible as we want the posteriors in the hidden layer to be certain about the conditioned input sample. The hyper-parameter that will control this is the softmax gain $\gamma$. We computed the conditional entropy of all HCs per sample, and plotted the histogram over all samples in Figure 3d. The histogram shows that the entropy predominantly has values $< 2$, whereas the maximum entropy is around $log(p_{MaxEnt}) \approx 4.6$ for $\gamma = 1$. This confirms that the bias regulation does not force the representations to have high marginal entropy at the cost of making all posterior per sample have high entropy. Figure 3b shows an interesting relationship between accuracy of the representations and the softmax gain $\gamma$. One would expect low values of $\gamma$ to have poor performance since we "flatten" the posteriors to be equal in value, and thereby, losing information about the input sample. However, high values of $\gamma$ (>1) also worsen the performance, that is, having "winner-take-all" like representations need not necessarily imply better representations.

### B. Structural plasticity

In Figure 4, we visualize the receptive fields of four randomly chosen HCs and a subset of the corresponding MCs in the network. Notice that we obtain rather contiguous patches even though no spatial structure of images were presented. The receptive fields of MCs also seem to capture diverse features such as lines and strokes.

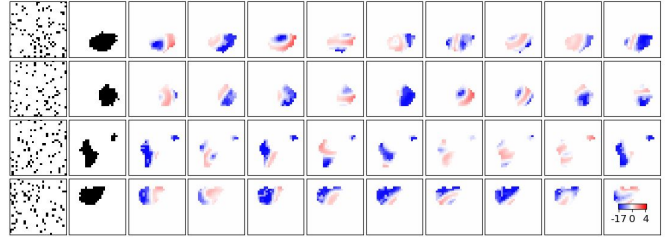

Fig 4: Receptive fields. Each row corresponds to a randomly chosen HC and the constituent MCs.. First column shows the receptive field of HC before training and second column after training (black means $M_{ij} = 1$). The remaining columns shows the receptive field of nine randomly chosen MCs in the HC.

The parameter $N_{flip}$ was introduced to control the number of flips in the receptive field per iteration. We measure the accuracy while varying $N_{flip}$ from 1 to 258 exponentially in steps of 2. Figure 5 shows that the accuracy converges at $N_{flip} = 16$.

### C. Classification

Table 2 shows the accuracy when learning with the following strategies: (i) Go, (ii) No-Go and, (iii) Go + No-Go. Go and No-Go strategies perform well individually, but Go + No-Go performs slightly better.

For the parameter set we found best suited (Table 1), we report the train and test accuracies as $99.10 \pm 0.63$ % and $96.49 \pm 0.12\%$, respectively.

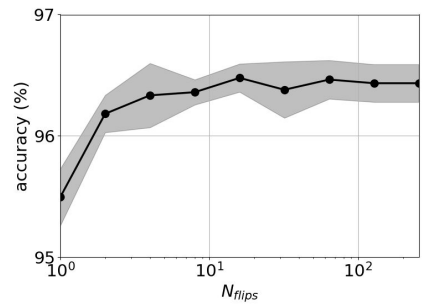

Figure 5: Accuracy results as a function of number of receptive field flips per iteration

TABLE II. CLASSIFICATION RESULTS

| Architecture | Accuracy (train) | Accuracy (validation) |
|---|---|---|
| Go | 98.03 ± 0.28 | 96.40 ± 0.10 |
| No-Go | 97.21 ± 0.13 | 96.23 ± 0.13 |
| Go + No-Go | 99.10 ± 0.63 | 96.52 ± 0.08 |

## VII. Discussion

We have demonstrated that the proposed network model can perform unsupervised representation learning using local Hebbian rules. The performance on MNIST is significantly lower than the "superhuman" deep learning methods. However, we consider it to be of lesser importance than the lower complexity of our correlation based brain-like learning approach, which has a potential for high robustness, good scaling and low-power hardware implementations.

It is important to note that the unsupervised learning methods introduced here are proof-of-concept designs and not meant to directly model some specific biological system or structure. Yet, they may shed some light on the hierarchical functional organization of e.g. sensory processing streams in the brain.

Further work will focus on extending our architecture with a brain-like deep structure and recurrent connectivity as well as to compare functionality and performance to other popular unsupervised learning networks such as stacked auto-encoders and stacked RBMs [2], as well as the network by Krotov and Hopfield [14].

### Acknowledgment

This project is funded within the framework of Swedish e-Science Research Centre (SeRC). The simulations were performed on resources provided by the Swedish National Infrastructure for Computing (SNIC) at the PDC Center for High Performance Computing, KTH Royal Institute of Technology.